\pdfoutput=1

\documentclass[11pt]{article}

\usepackage{ACL2023}
\usepackage{times}
\usepackage{latexsym}

\usepackage[T1]{fontenc}

\usepackage[utf8]{inputenc}

\usepackage{microtype}

\usepackage{amsmath}
\usepackage{array}
\usepackage{xspace}
\usepackage{arydshln}
\newcommand{\dataset}{\textsc{OverMiss}\xspace}
\newcommand{\model}{\textsc{SWIE}\xspace}
\usepackage{inconsolata}
\usepackage{booktabs}
\usepackage{xcolor}
\usepackage{colortbl}
\usepackage{rotating}
\usepackage{multirow}

\title{Improving Translation Faithfulness of Large Language Models via Augmenting Instructions}

\author{Yijie Chen$^1$, Yijin Liu$^2$, Fandong Meng$^2$, Yufeng Chen$^1$, Jinan Xu$^1$, Jie Zhou$^2$ \\
  $^1$Beijing Jiaotong University, Beijing, China\\
  $^2$Pattern Recognition Center, WeChat AI, Tencent Inc, China \\
    {\tt \{22120354, chenyf, jaxu\}@bjtu.edu.cn}\\
    {\tt \{yijinliu, fandongmeng, withtomzhou\}@tencent.com}
 }

\begin{document}
\maketitle
\begin{abstract}
Large Language Models (LLMs) present strong general capabilities, and a current compelling challenge is stimulating their specialized capabilities, such as machine translation, through low-cost instruction tuning. 
The standard instruction-following data is sequentially organized as the concatenation of an instruction, an input, and a response. 
As the attention mechanism of LLMs has limitations on local focus, LLMs tend to focus more on the words or sentences nearby at each position. This leads to a high risk of instruction forgetting during decoding. 
To alleviate the above issues, We propose {\model} (\textbf{S}egment-\textbf{W}eighted \textbf{I}nstruction \textbf{E}mbedding) and an instruction-following dataset {\dataset}. {\model} improves the model instruction understanding by adding a global instruction representation on the following input and response representations. 
\dataset improves model faithfulness by comparing over-translation and miss-translation results with the correct translation.
We apply our methods to two mainstream open-source LLMs, BLOOM and LLaMA. 
The experimental results demonstrate significant improvements in translation performance with SWIE based on BLOOMZ-3b, particularly in zero-shot and long text translations due to reduced instruction forgetting risk. 
Additionally, OVERMISS outperforms the baseline in translation performance ({\em e.g.} an increase in BLEU scores from 0.69 to 3.12 and an average improvement of 0.48 percentage comet scores for LLaMA-7b) with further enhancements seen in models combining OVERMISS and SWIE ({\em e.g.} the BLUE scores increase up to 0.56 from English to German across three different backbones), and both exhibit improvements in the faithfulness metric based on word alignment. \footnote{Our code and datasets are released in Github: https://github.com/pppa2019/swie\_overmiss\_llm4mt}
\end{abstract}

\section{Introduction}

In recent years, pre-trained language models (PLMs) have experienced a burgeoning growth and have been extensively investigated and employed in downstream tasks. However, large language models (LLMs) exhibit surprising emergent abilities~\cite{wei2022emergent} which have not been observed in small PLMs, and LLMs have shown significant ability on general tasks and zero-shot or few-shot settings, even including symbolic reasoning, commonsense, algorithm, and so on. \par

Super LLMs like GPT-4 and ChatGPT, which can only be used via API, have demonstrated remarkable translation performance without fine-tuning~\cite{jiao2023chatgpt,hendy2023good}. For general LMs, fine-tuning is a prevailing approach to adapt to specific downstream tasks. Consequently, the fine-tuning of relatively smaller open-source LLMs presents an attractive alternative, given that it can augment the model's translation capabilities without imposing significant computational costs~\cite{jiao2023parrot}. Nonetheless, instruction tuning on LLMs in machine translation remains a field that has not been fully explored. \par

Although the de facto architecture of state-of-the-art models in machine translation remains encoder-decoder~\cite{bahadanau2015neural,gao2022encoder}, the majority of the open-source LLMs adopt the causal language model (causal LM) architecture. 
However, the core limitation of causal LM is the local focus~\cite{liu2023lost} of its attention mechanism, which leads to the model's tendency to focus on nearby words or sentences at each position. Consequently, in the instruction fine-tuning data, the instruction text is further away from the output compared to the input text, increasing the risk of instruction forgetting during decoding. In machine translation, ignoring instructions can lead to issues such as hallucinations or unfaithfulness, ultimately reducing the quality and credibility of the models.

This paper introduces a novel method for improving instruction tuning named \model (\textbf{S}egment-\textbf{W}eighted \textbf{I}nstruction \textbf{E}mbedding), which utilizes parameterized adapters to encode instruction and introduces segmented weight to enable a natural integration of instruction representations and global representations. In order to further improve the model translation faithfulness, we present \dataset, an instruction dataset that utilizes our proposed framework to collect contrastive negative samples that specifically target over-translation and miss-translation issues. 

We evaluate our methods on different machine translation benchmarks and various backbone models (including BLOOMZ-3b, BLOOMZ-7b1-mt, and LLaMA-7b). Results on BLOOMZ-3b show \model has improved from 0.19 to 0.51 BLEU scores on four translation directions of WMT22 test sets, from 0.20 to 0.58 BLEU scores on six zero-shot translation directions, and 0.67 BLEU scores in average on long sentence test sets. 
Additionally, \dataset also leads to significant improvements ({\em e.g.} for WMT22 test sets, an increase in BLEU scores from 0.69 to 3.12 and an increase in 0.48 percentage COMET score on average on LLaMA-7b on WMT). The combination of \model and \dataset achieves a further improvement up to 0.56 BLEU scores on three backbone models from English to German.
\par
In summary, our contributions are as follows:
\begin{itemize}
    \item We propose \model, a novel segment-weighted instruction embedding method, which effectively improves translation performance and faithfulness, and its effectiveness is more significant in the zero-shot and longer text settings since the strengthening of instruction-following ability.
    \item We propose a translation faithfulness contrastive instruction-tuning data construction method and construct \dataset. We demonstrate that \dataset consistently improves the translation performance on three backbone models and two test sets ({\em e.g.} on LLaMA-7b, increase up to 3.12 BLEU score on WMT22 test sets and up to 3.03 BLEU score on FLORES test sets.)
    \item By examining the internal attention scores of the models, we discovered that \model leads to a higher attention ratio for instructions compared with baseline, thereby validating our hypothesis and effectively substantiating its efficacy in mitigating the instruction forgetting problem.

\end{itemize}

\begin{figure}[ht]
    \centering
    \includegraphics[width=0.35\textwidth]{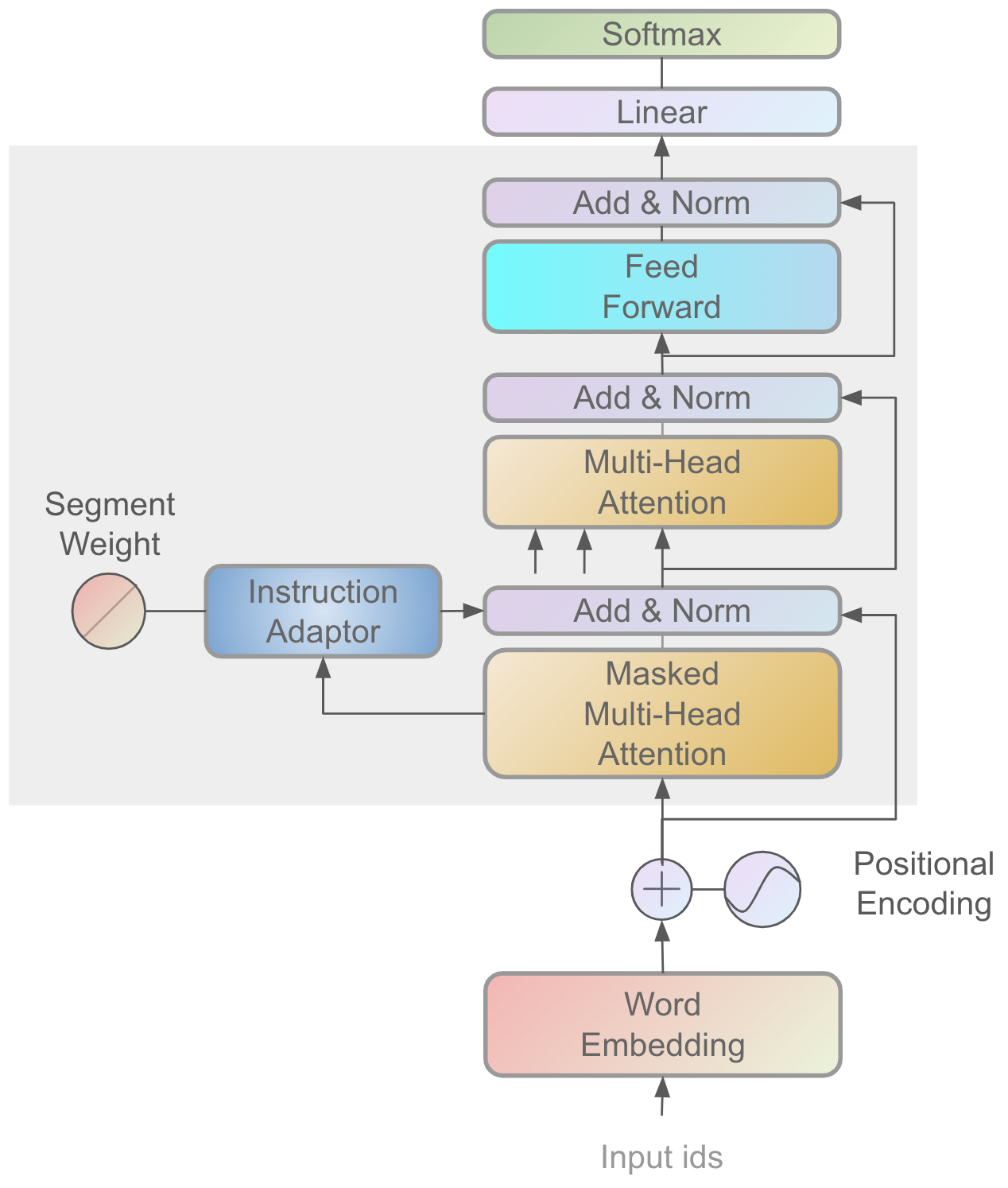}
    \caption{The model structure of \model.}
    \label{fig:model}
\end{figure}
\begin{figure*}
    \centering
    \includegraphics[width=0.65\textwidth]{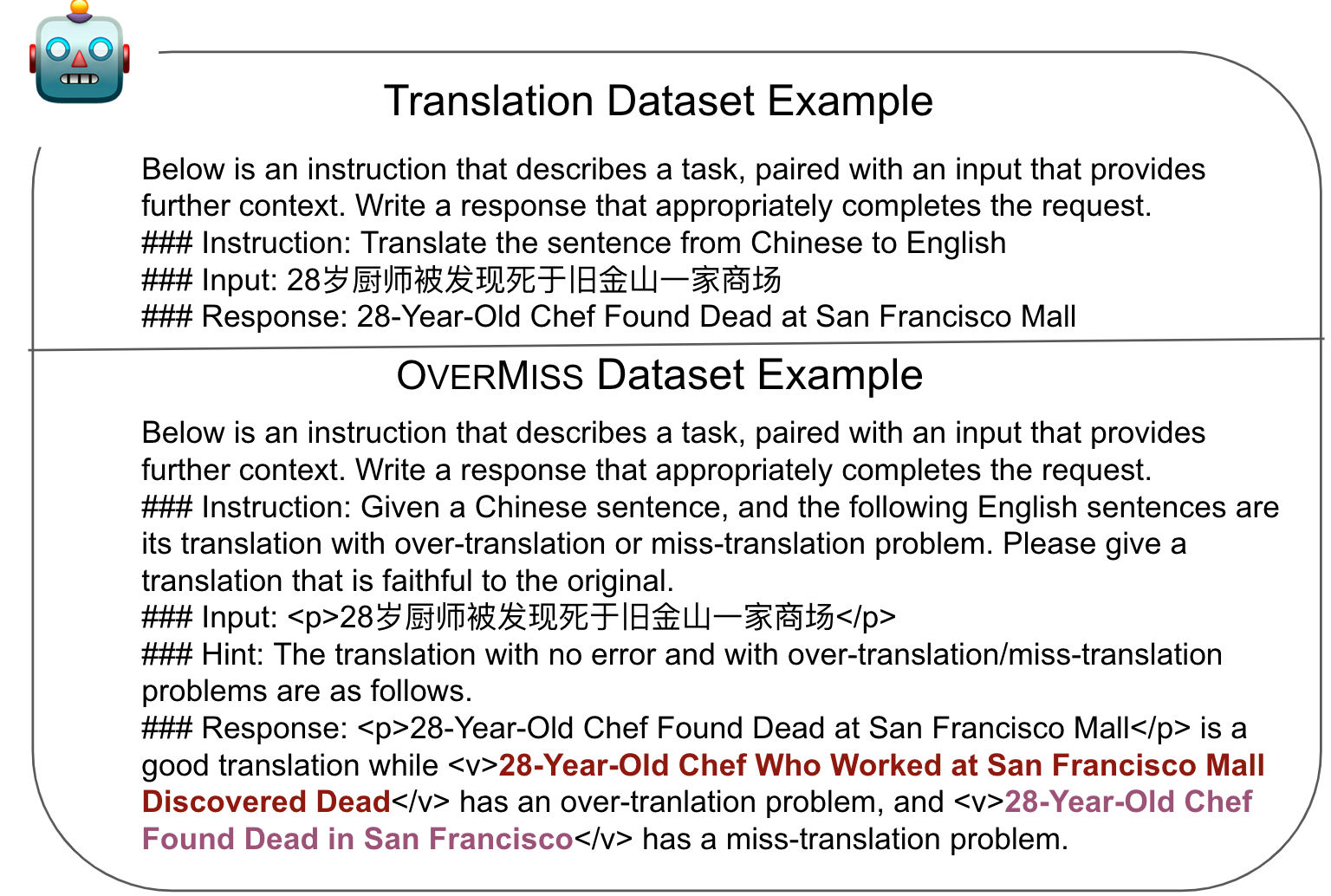}
    \caption{An instance of translation instruction and an instance of \dataset.}
    \label{fig:data_example}
\end{figure*}

\section{Related work}
Our work is closely related to machine translation, the variants of instruction tuning for LLMs, and hallucination in text generation. We will provide a brief overview of these areas in this section.
\subsection{Machine Translation based on LLMs}
Owing to the strong zero-shot and instruction-following abilities of LLMs, super LLMs like GPT-4 have achieved comparable translation performance to the best system on WMT system in high-resources translation direction on translation and relevant tasks like post-editing~\cite{raunak2023leveraging,he2023exploring}.

The aforementioned study exclusively employs models that are only accessed via API, thereby limiting its applicability. Consequently, numerous studies have been conducted to investigate the potential of fine-tuning open-source LLMs.
In the context of instruction tuning LLMs for machine translation, \cite{jiao2023parrot,zhang2023bayling} have proposed multi-task instruction data construction frameworks for instruction tuning open-source LLMs on machine translation. \cite{zeng2023tim} proposed a contrastive learning loss in order to train the model to learn contrastive sample pairs.\par

\subsection{Instruction Tuning}
The first work on instruction tuning is FLAN~\cite{wei2021finetuned}, which shows a surprising result on zero-shot and few-shot settings. There are a lot of follow-up works proposed to construct larger-scale instruction datasets. The instruction tuning datasets adopt different instruction and language styles: FLAN~\cite{longpre2023flan} use ``input'' and ``target''; unnatural instruction\cite{honovich2022unnatural} use ``instruction'', ``input'', ``constrain'' and ``output''; Super-NaturalInstructions\cite{wang2022super} constructs positive and negative sample for each task. As a unified and scaling-up dataset, OPT-IML casts all the above datasets to ``instruction'' and ``output'' segments. \par
Due to the fact that instructions serve as the definetion of tasks and are typically located at the beginning of samples, the representation of instructions in causal LMs face a higher risk of being forggoten during decoding.  To allevate this issue, there are currently some efforts proposing improved methods that differ from the standard fine-tuning approaches, in order to enhance the learning in instruction components.


\cite{ye2022guess} models the instruction in the condition given input and target, thereby alleviating the demands of long context modeling. \cite{choi2022prompt} proposed a distilling-based context injection method to preserve the long context information in the fixed model when the model is used in static long prompts situations. 
\par
As the above methods require higher demands for data and task scenarios, such as fixed instructions as conditions. They cannot meet the condition of machine translation, which typically only contains short instructions that indicate the translation direction.

\subsection{Hallucinations in Language Models}
Hallucinations in neural machine translation have been discussed for a long time~\cite{leehallucinations,muller-2020-domain}, and it has the same mean as unfaithfulness. It is widely observed that the sources of hallucination or unfaithfulness can be the lack of knowledge or inadequate attention to the source ~\cite{ferrando2022towards,raunak2021curious}. 

On machine translation hallucination detection benchmarks, we found that existing datasets are constructed by humans or perturbing the translation model~\cite{raunak2021curious}. Human-making datasets like HalOmi~\cite{dale2023halomi} are highly cost and hard to scale up. Datasets generated by the model perturbing method have low quality because the sentences generated are far from both the natural distribution and the distribution of modern LLMs. Thus, our proposed hallucination-mimicking dataset construction method can fill the gap with high-quality fluent negative samples. \par

\section{Method}
In this section, we propose a contrastive faithfulness translation instruction dataset \dataset and a global instruction fusion method \model. Before introducing the proposed method, the necessary background will be formulated first. 

\subsection{Backgound}
\subsubsection{Instruction Tuning Formalization}
Instruction tuning is one of the alignment methods to make language models meet human preferences. To formulate instruction tuning, we define $s$, $x$, and $y$ as the instruction, the input, and the target, respectively. Noting that the input is not necessary but the instruction is needed all the time. The standard instruction tuning is trained with maximum likelihood estimation (MLE), and the training objection can be calculated by Equation (\ref{eq:loss}). Furthermore, due to the attention mechanism tends to pay more attention to the text nearby, the instruction part faces a higher risk to be forgotten in the generation process.
\begin{equation}
    L_{MLE} = -\sum_{t=1}^{T}{\rm log}P(y_t \mid y_{<t}; x; s)
    \label{eq:loss}
\end{equation}

\subsubsection{Causal Language Model}

Decoder-only architecture is designed for unified text generation tasks, including prefix decoder and causal decoder~\cite{raffel2020exploring}. Most of the LLMs use the causal decoder architecture because of the wide observation of scaling law on the causal decoder. However, a more comprehensive investigation of other architectures' performance at a large scale is still lacking.~\cite{zhao2023survey}\par
A causal language model is composed of a stack of causal decoder layers.
The function of the multi-head attention mechanism is to combine the hidden representation of each position with contextual information. With a causal attention mask, text generation tasks in any format can be unified in training and decoding states. In detail, let $m$, $n$ be the position indexes of two tokens, and $\boldsymbol{q}$, $\boldsymbol{k}$, $\boldsymbol{v}$, $\boldsymbol{o}$ are, respectively, query, key, value, and output representation, and the length of input tokens is $N$. In a causal language model, when $m > n$, the attention score $a_{m,n}$ will be masked.
\begin{equation}
    \boldsymbol{a}_{m, n}=\frac{\exp \left(\frac{\boldsymbol{q}_m^{\top} \boldsymbol{k}_n}{\sqrt{d}}\right)}{\sum_{j=1}^N \exp \left(\frac{\boldsymbol{q}_m^{\top} \boldsymbol{k}_j}{\sqrt{d}}\right)}
\end{equation}
\begin{equation}
    \boldsymbol{o}_m = \sum_{n=1}^{N} \boldsymbol{a}_{m,n} \boldsymbol{v}_{n}
\end{equation}

\subsection{\model: Segment-weighted Instruction Embedding}
We propose segment-weighted instruction embedding in order to strengthen global instruction attention for decoders, and the details are described as follows.
Instruction tuning can be divided into several segments, including instruction, input, response, and so on.
The sentence will be converted to a list of tokens after the tokenizer. We define a segment ID for each segment and then map the segment index to every token of the tokens list.
Assuming a sentence tokens list is represented as $\mathbf{S}=\left[\mathbf{s}_1, \mathbf{s}_2, \cdots, \mathbf{s}_l\right] \in \mathbf{R}^{1 \times N}$, and its segment ids list is $I_{s}$, which is also an array with length $N$. Let the $c$ be the ID of the instruction span, and the $b$ be the array record of the beginning indexes of each span. The encoded instruction representation can be obtained in the output of each decoder layer, we use an instruction adapter to re-parameterize instruction. 
We set a segment weight to constrain the fusion of instruction representation on input and response segments.
Let $L$ be the length of the tokens list of the input span, $B$ be the array that records the beginning position index of each segment, and we use $L$ to standardize the slope. The $H_l$ represents the hidden output of $l^{th}$ layer and the $H_{ins_l}$ represents the max pool result of the instruction part in $H_l$. We use a down-sampling linear layer, an activation layer, and an up-sampling linear layer as the adapter. \par
\begin{figure}
    \centering
    \includegraphics[width=0.9\linewidth]{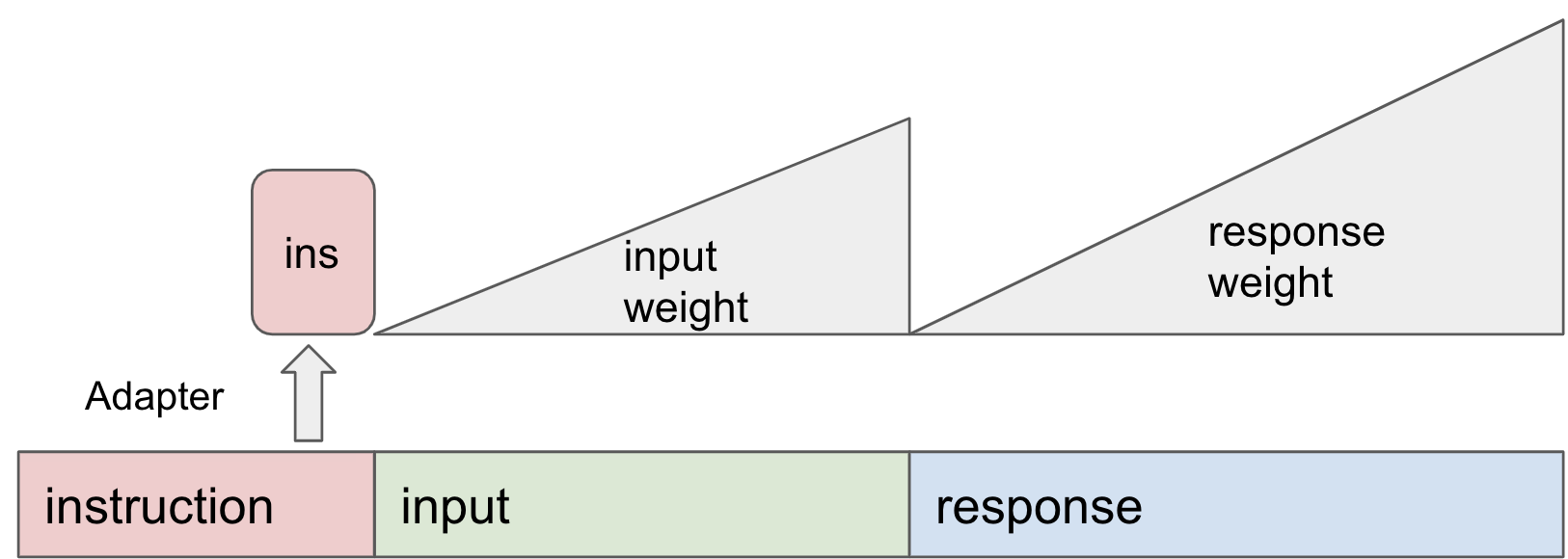}
    \caption{An illustration of segmented-weight}
    \label{fig:weight}
\end{figure}

On implementation details, we selected the middle three layers of the language model to fuse the extracted instruction feature with the global hidden representation. The selection principle is based on our analysis of the attention score distribution of each layer, and the detailed analysis process is shown in Section. \textit{Visualize Inadequate Attention on Instruction}.
The model structure is visualized in Figure.\ref{fig:model}, and the process is described as Equation~(\ref{eq:overall}-\ref{eq:adaptor}), and the illustration of segmented-weight is shown in Figure.\ref{fig:weight}.
\begin{equation}
    H_l := H_l + W_{seg} \cdot f(H_{ins_l})
    \label{eq:overall}
\end{equation}

\begin{equation}
W_{seg_{i}}=\left\{
\begin{array}{rcl}
0 & & {I_{s}[i] = c}\\
\frac{i-B[I_s[i]]}{L}& & {I_s[i] \neq c}\\
\end{array} \right.
\label{eq:weight}
\end{equation}

\begin{equation}
    f(H_{ins_l}) = \mathbf{L_{up}}(\sigma(\mathbf{L_{down}}(H_{ins_l})))
    \label{eq:adaptor}
\end{equation}


\subsection{{\dataset}: A Natural Hallucination Dataset}\label{sec:data_pipeline}
In the machine translation task, the most usual taxonomy of model hallucination or unfaithfulness for fluent output is over-translation and miss-translation. Over-translation refers to the situation in which the translated sentence contains words irrelevant to the source sentence, and miss-translation refers to the situation in which the translation sentence lacks part of the information from the source sentence. 
Thus, we prompt gpt-3.5-torbo to mimic the two typical error types, and the prompts are appended in Table.\ref{tab:halomi_prompt}.\par
To qualify the extent of miss-translation or over-translation errors of generated sentences, we use awesome-align\footnote{https://github.com/neulab/awesome-align} to evaluate the word-level cross-lingual alignment rate, and the statistic result is shown in Table.\ref{tab:data_stat}, which indicates the generated data satisfied the requirement of the negative samples while preserving the meaning of most of the source sentences.\par
Instruction-tuning datasets can be organized flexibly, and the standard format contains instruction, input, and response. After we constructed the over-translation and miss-translation contrastive samples based on WMT17-20 with the proposed automatic pipeline, we organized the final instruction data as Figure.\ref{fig:data_example}. And the total number of samples in the dataset is 54420.

\begin{table*}
\centering
\resizebox{0.85\linewidth}{!}{
    \begin{tabular}{lp{12cm}}
    \toprule
    type & prompt\\
    \midrule
    miss-translation& You are an unprofessional [source language] to [target language] translator who is not fully faithful to the original text in the translation process there is a problem of omission, i.e. the translation leaves out parts of the original text.\newline Please translate the following [source language] sentence:\newline [source sentence]\newline If the following is a high-quality human [target language] translation:\newline [target sentence]\newline Please give a direct low-quality [target language] translation with omission problems, noting that you are not simply rewriting the previous translation, but need to emulate a translator that may have omissions, i.e. omitting parts of the original text. \\
    \hdashline
    over-translation&  You are an [source language] to [target language] translator, but your translation is not professional. In the translation process, you have not been completely faithful to the original text, resulting in a translation that is not in the original text. \newline
    This is a translation illusion problem and you need to give a translation that has the illusion problem. Please translate the following [source language] sentence:\newline[source sentence]\newline 
    If the following is a high-quality human [target language]translation:\newline[target sentence]\newline
    Please give a straightforward low-quality [target language] translation that has an additive translation problem or a translation illusion problem. Please note that you need to simulate a translator with possible translation enhancement problems and translate what is not in the original text, rather than simply rewriting the previous translation.\\
    \bottomrule
    \end{tabular}
    }
    \caption{The prompts for producing the \dataset dataset.}
    \label{tab:halomi_prompt}
\end{table*}

\begin{table}[]
    \centering
    \begin{tabular}{lcc}
\toprule
data & source coverage & target coverage\\
\midrule
reference&0.8845&0.8699\\
miss\ data&0.5800&0.7180\\
over\ data&0.6958&0.5771\\
\bottomrule
    \end{tabular}
    \caption{Data statistics of generated over-translation and miss-translation data.}
    \label{tab:data_stat}
\end{table}


\section{Emprical Experiments}
We choose BLOOM and LLaMA as the backbone models. There are 4 translation directions included, De $\Rightarrow$ En, En $\Rightarrow$ De, En $\Rightarrow$ Zh, and Zh $\Rightarrow$ En.
\subsection{Training Setting}

\subsubsection{Alpaca}
Alpaca Dataset\footnote{https://github.com/tatsu-lab/stanford\_alpaca} is a high-quality multi-task instruction-following dataset that contains 52K items. We use Alpaca Dataset to finetune the pre-trained language models as our baseline.
\subsubsection{Parrot-hint}
Following ~\cite{jiao2023parrot}, we set Parrot-hint as our strong baseline. The Parrot-hint\footnote{https://github.com/wxjiao/ParroT} dataset includes 3 sub-datasets, Alpaca Dataset, the WMT17-20 dataset, and the MQM instruction dataset. Parrot-hint contains 200K data in total.
\subsubsection{\dataset}

In the training process, we utilize the Parrot-hint dataset to ensure the basic ability of the fine-tuned models. As the mixup dataset contains instruction-following data without a hint and with a hint, and data with a hint both have an auxiliary task based on translation. So we use a curriculum learning strategy to fine-tune the data in two stages.

\subsection{Evaluation}
This section introduces the test sets and the evaluation metrics we use. 
\subsubsection{WMT22 Test Sets}
WMT22 test sets come from the news translation track of WMT22 competition\footnote{https://github.com/wmt-conference/wmt22-news-systems}. The test sets include 1984, 2037, 2037, and 1875 samples for De $\Rightarrow$ En, En $\Rightarrow$ De, En $\Rightarrow$ Zh, and Zh $\Rightarrow$ En, respectively.
\subsubsection{Flores-200 Dev-test}
Flores-200 is a multi-language translation benchmark. We use the dev-test split as our test set to enrich our experiments, and there are 1012 samples for each translation direction.
\subsubsection{Automatic Evaluation}
For lexical evaluation, we use BLEU~\cite{papineni2002bleu}; for semantic evaluation, we use COMET with reference. Both of them are widely used metrics in machine translation, and we use ScareBLEU\footnote{https://github.com/mjpost/sacrebleu} and Unbabel/wmt22-comet-da in the evaluation implementation.

\subsection{Implement details}
We use the transformers and DeepSpeed framework for model training and inference.
The training hyper-parameters follow the setting of~\cite{jiao2023parrot}.
We uniformly set the dim of the instruction adapter to 32.
The 3B size models are trained on 8 V100 GPUs, and the 7B size models are trained on 4 A100(40G) GPUs. In order to reduce the memory requirement and prevent the models from over-fitting, we train all models with freezing embedding layers in DeepSpeed stage 1.

\subsection{Main Results}
The main results are shown in Table.\ref{tab:main_res}. 
For LLaMA fine-tuned by Alpaca, the model performs well when translating En$\Leftrightarrow$De, while when translating En $\Leftrightarrow$ Zh, it often confuses the target language, resulting in code-mixing or out-of-target translation. For BLOOM fine-tuned by Alpaca, the model translates En $\Leftrightarrow$ Zh better while translating worse in En$\Leftrightarrow$De when comparing with LLaMA-Alpaca, indicating the difference between the basic language translation capacity of models.
Overall, we have three main observations during the experiment as follows.\par
Firstly, according to the comparison between \dataset and Parrot-hint, we found that \dataset notably led to performance enhancement. For example, based on LLaMA-7b, the model trained with \dataset has an improvement of 1.02, 1.25. 3.12 and 0.69 BLEU scores on four translation directions respectively, and an improvement of 0.48 percentage comet scores on average. 
Although the Flores dataset has a different distribution from the WMT training data, we found that the \dataset still increases 0.46 BLEU score on En $\Rightarrow$ De and 3.03 BLEU score on Zh $\Rightarrow$ En.\par
Secondly, according to the comparison between \model and Parrot-hint, we found that this method has an obvious improvement on some of the settings and has a stable slight improvement on other settings. For example, on BLOOMZ-3b, \model outperforms Parrot-hint from 0.19 to 0.51 BLEU scores.\par
Thirdly, by combining the \dataset and \model, a further improvement can be seen in all of the backbones in the En $\Rightarrow$ De translation direction from 0.05 to 0.56 BLEU scores, and in BLOOMZ-7b-mt in three of the four translation directions. Since both methods aim to improve faithfulness, their combination is not orthogonal.

\begin{table*}[h]
    \centering
    \scalebox{0.9}{
    \begin{tabular}{lrrrrrrrr}
\toprule
\multirow{2}{*}{Model}  & \multicolumn{2}{c}{De $\Rightarrow$ En } & \multicolumn{2}{c}{En $\Rightarrow$ De} & \multicolumn{2}{c}{En $\Rightarrow $ Zh} & \multicolumn{2}{c}{Zh $\Rightarrow$ En} \\
\cmidrule(rl){2-3} \cmidrule(rl){4-5} \cmidrule(rl){6-7} \cmidrule(rl){8-9}
\multicolumn{1}{r}{} & \multicolumn{1}{l}{bleu} & \multicolumn{1}{l}{comet} & \multicolumn{1}{l}{bleu} & \multicolumn{1}{l}{comet} & \multicolumn{1}{l}{bleu} & \multicolumn{1}{l}{comet} & \multicolumn{1}{l}{bleu} & \multicolumn{1}{l}{comet} \\
\midrule
\multicolumn{9}{c}{WMT22 Winners}\\
&33.7  & 85.46      & 38.4  & 88.09      & 33.5  & 87.84      & 54.3  & 81.12        \\
\hline
\multicolumn{9}{c}{BLOOMZ-3b WMT22} \\
Alpaca & 14.68 & 68.49 & 5.55  & 49.10  & 20.20  & 81.46  & 11.65 & 75.38 \\
Parrot& 22.05 & 75.59 & 17.80  & 67.64 & 33.95 & 83.70  & 21.33 & 78.19 \\
w/ \model &22.56 & 75.59 & 18.17 & 67.64 & 34.14 & 83.64 & 21.71 & 78.58  \\
w/ \dataset &24.00    & \textbf{76.66} & \textbf{19.24} & \textbf{70.44} & 35.35 & \textbf{83.51} & \textbf{21.93} &\textbf{78.08} \\
w/ \dataset w/ \model & \textbf{24.05} & 76.40  & 19.03 & 70.38 & \textbf{35.48} & 83.34 & 21.73 & 78.06  \\
\hline
\multicolumn{9}{c}{BLOOMZ-7b1-mt WMT22} \\

Alpaca & 18.64 & 73.37 & 9.97 & 61.65 & 25.52 & 82.31 & 15.07 & 77.79 \\
Parrot & 23.80  & 77.77 & 20.58 & 73.63 & 35.49 & 84.61 & 22.58 & 78.93  \\
w/ \model & 24.34 & 77.90  & 20.19 & 73.17 & 35.99 & \textbf{85.02} & 22.28 & 79.22  \\
w/ \dataset &    25.84 & 78.79 & \textbf{22.15} & 75.01 & 36.61 & 84.43 & \textbf{23.40}  & \textbf{79.36}   \\
w/ \dataset w/ \model &   \textbf{25.95} & \textbf{78.80}  & 21.83 & \textbf{75.17} & \textbf{36.88} & 84.53 & 23.33 & 79.15  \\

\hline
\multicolumn{9}{c}{LLaMA-7b WMT22} \\
Alpaca &   28.92 &   82.77     & 21.72 &   79.70     & 17.72 &     71.96   & 15.95 &  74.95    \\

Parrot &  28.90  & 82.84 & 25.96 & 82.78 & 28.12 & 79.84 & 20.61 & 75.61 \\
w/ \model &   28.56 & 82.97 & 25.70  & 82.11 & 29.03 & 79.68 & 20.33 & 75.48   \\
w/ \dataset &  29.92 & \textbf{83.50}  &\textbf{ 27.21} & \textbf{82.36} & \textbf{31.24} & \textbf{80.63} & \textbf{21.30}  & \textbf{76.48}  \\
w/ \dataset w/ \model &  \textbf{30.48} & 82.97 & 27.10  & 81.89 & 31.08 & 80.14 & 21.19 & 76.14  \\
\hline
\multicolumn{9}{c}{LLaMA-7b Flores} \\
Parrot-hint & 40.83 & 88.50  & 31.14 & 85.73 & 26.96 & 80.08 & \textbf{22.48} &\textbf{83.62} \\
w/ \model & \textbf{40.92} & 88.51 & 30.82 & 85.52 & 27.34 & 79.86 & 22.23 & 83.44 \\
w/ \dataset & 40.35 & \textbf{88.55} & \textbf{31.60} & \textbf{85.59} & \textbf{29.99} & \textbf{81.95} & 21.68 & 83.64 \\
w/ \dataset w/ \model &  40.20  & 88.39 & 31.41 & 85.21 & 29.07 & 81.14 & 21.59 & 83.50  \\
\bottomrule
\end{tabular}%
}
    \caption{Translation performance of LLMs on WMT22 and Flores test sets. The \textbf{bolded} scores refer to the best performance under the same or comparable settings.}
    \label{tab:main_res}
\end{table*}

\begin{table}[ht]
    \centering
    \resizebox{1\linewidth}{!}{
    \begin{tabular}{lllll}
    \toprule
        model setting & De $\Rightarrow$ En  & {En $\Rightarrow$ De} & En $\Rightarrow$ Zh & {Zh $\Rightarrow$ En} \\
    \midrule
        Parrot-hint &23.73 & 17.11 & 34.70  & 19.11 \\
        w/ \model & 24.02 & 17.79 & 34.94 & 20.59 \\
    \bottomrule
    \end{tabular}
    }
    \caption{The comparison between baseline and \model on WMT22-concat dataset.}
    \label{tab:concat}
\end{table}
\subsection{Long Sentence Translation}
To assess the efficacy of \model in the context of long text translation, we employed a concatenation approach to merge the adjacent 3-5 sentences from the WMT22 test sets, thereby creating the WMT22-concat test set for long text translation.
Subsequently, we conducted an ablation experiment on \model using the WMT22-concat test set on BLOOMZ-3b. The results are presented in Table.\ref{tab:concat} demonstrate that \model yields an average improvement of 0.6725 BLEU scores, with a notable increase of 1.49 BLEU score observed in the Chinese to English translation. These findings suggest that instruction augmenting is better suited to long text scenarios and can lead to further enhancements in performance compared to the original WMT22 test sets.





\subsection{Zero-shot Performance}
Using zero-shot translation directions, the instruction-following ability can be effectively evaluated.
We select 6 zero-shot directions from WMT22 test sets, including Uk $\Rightarrow$ En, Fr $\Rightarrow$ De, Cs $\Leftrightarrow$ En, and Ru $\Leftrightarrow$ En. We observe that \model leads to 0.20 to 0.58 BLEU scores. The experiment results are as our expectation since the \model enhances the instruction-following ability of the model, and the instruction needs more attention in the zero-shot translation direction scenario.
\begin{table*}[h]
    \centering
    \resizebox{0.8\linewidth}{!}{
    \begin{tabular}{lrrrrrr}
    \toprule
        setting & Uk $\Rightarrow$ En & Fr $ \Rightarrow $ De & Cs $ \Rightarrow$ En & En $\Rightarrow $ Cs  & Ru $\Rightarrow $En & En $\Rightarrow$ Ru \\
    \midrule
        Parrot-hint &6.77  & 19.58 & 4.97  & 2.69  & 17.59 & 4.35  \\
        w/ \model & 7.33  & 19.73 & 5.14  & 2.85  & 17.79 & 4.74  \\
    \bottomrule
    \end{tabular}
    }
    \caption{Zero-shot BLEU scores performance based on BLOOMZ-3b.}
    \label{tab:zero_shot}
\end{table*}
\subsection{The Impact of Inference Instruction}
We test the impact of inference prompts. As the auxiliary task instruction dataset provides the model with typical translation quality information, we can use more detailed prefixes during the inference stage to guide the model's translation process with an awareness of certain principles.
In Table.\ref{tab:infer_prompt}, the basic setting means the briefest instruction, that is, ``translate the following sentences from [source language] to [target language]". According to the training task contained in the datasets, we use some extra guides to provide a more detailed request to models, such as output with no error or no over/miss-translation problems. Conflicting to the findings in ~\cite{jiao2023parrot}, the ``no-error" hint does not yield positive benefits in the situation where fine-tuned model on \dataset, while the ``no-over", ``no-miss" and ``no-over/miss" can improve model performance furthermore.
\begin{table}[ht!]
    \centering
    \resizebox{0.7\linewidth}{!}{
    \begin{tabular}{lrr}
\toprule
setting & Overall BLEU\\
\midrule
basic& 25.13 \\
w/ no-error & 24.76   \\
w/ no-over & 25.13   \\
w/ no-miss & \textbf{25.29} \\
w/ no-over/miss & 25.24  \\
\bottomrule
\end{tabular}
}
    \caption{The comparison between inference prompts.}
    \label{tab:infer_prompt}
\end{table}

\subsection{Faithfulness Quantification}
On the qualification of word-level machine translation faithfulness, there is no widely-used standard toolkit yet. 
The same method as Section.\textit{Natural Hallucination Data Construction}, we use word alignment tools to match the source sentences and the inference sentences word by word, then calculate the recall of source words matching rate and hypothesis words matching rate, and then the ratio can reflect the absence and the redundancy extent. The final scores are derived by averaging the source and target coverage rate on our WMT22 test sets.
The result shows in Table.\ref{tab:faithful} that both {\model} and {\dataset} can improve the faithfulness of results, showing the effectiveness of our proposed method.
\begin{table}[h]
    \centering
    \resizebox{0.7\linewidth}{!}{
    \begin{tabular}{ll}
    \toprule
        setting & score \\
    \midrule
        Parrot-hint & 87.94 \\
        w/ \model & 88.28 \\
        w/ \dataset &\textbf{88.84} \\
        w/ \model w/ \dataset & 88.80 \\
    \bottomrule
    \end{tabular}
    }
    \caption{The ablation study of faithfulness score on \model and \dataset.}
    \label{tab:faithful}
\end{table}

\section{Visualize Inadequate Attention on Instruction}\label{sec:attn_anaylsis}
Our standard instruction-following data item is organized as instruction, input, and output sequentially. The attention score in transformers can show the positions the model addresses more. We divide a random translation sample from test sets into 3 spans, including instruction, input, and response. Subsequently, we calculate the accumulation attention scores for each span on each token. Assuming $a$ is the attention score matrix, the $sid$ is the index of the end of the span, we use $S_{span}$ to represent the accumulated attention score in a position as shown in Equation.\ref{eq:accu}.\par
\begin{figure}[h]
    \centering
    \includegraphics[width=0.9\linewidth]{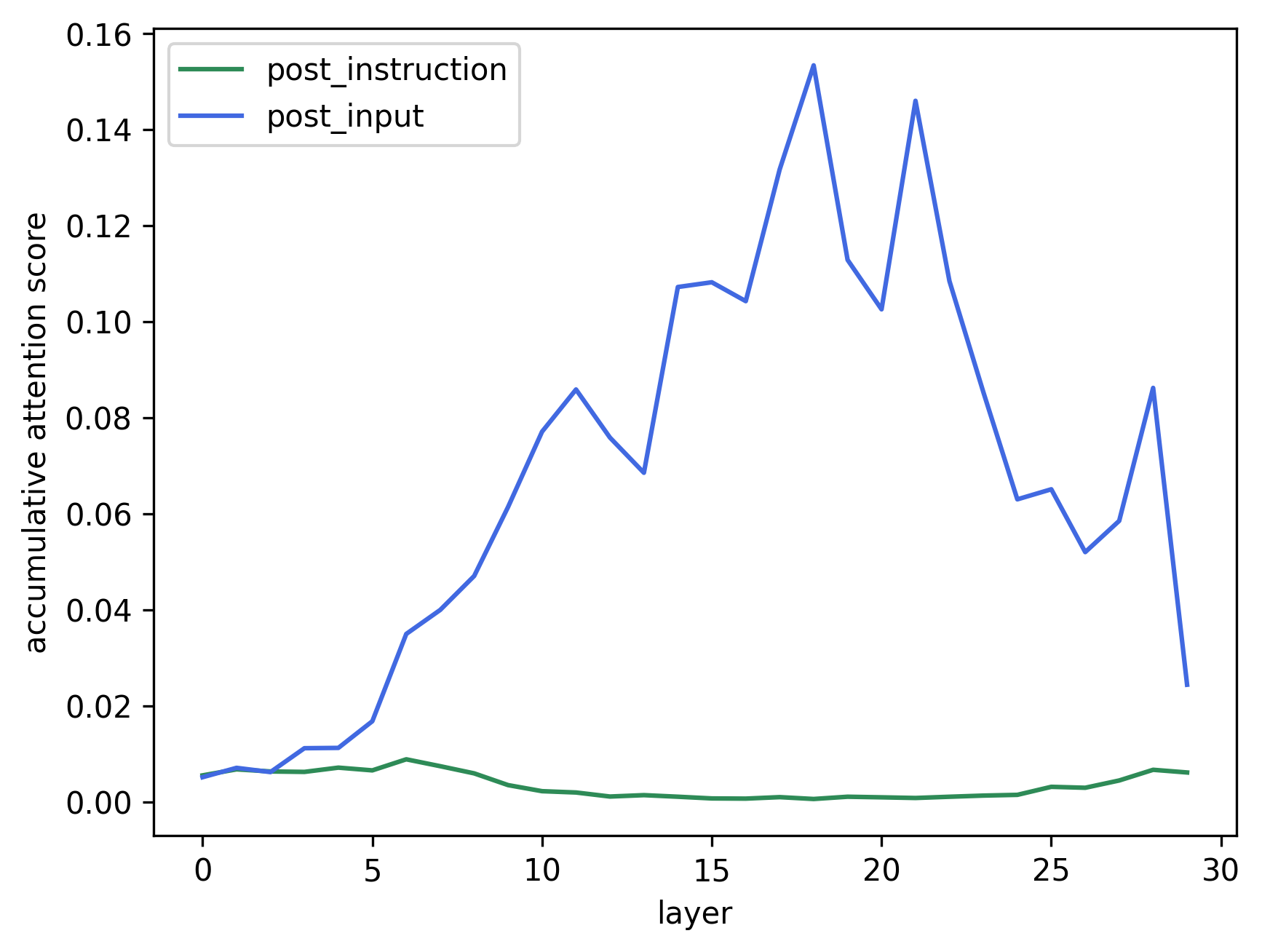}
    \caption{Accumulative attention score on the post-instruction and post-input positions for each layer. This figure is based on BLOOMZ-3b fine-tuned by the Parrot-hint dataset in the origin model structure.}
    \label{fig:attn-layer}
\end{figure}

As depicted in Figure.\ref{fig:attn-layer}, it is evident that the middle layers of the model manifest a considerably higher attention accumulation score on the input spans, whereas the bottom and top layers exhibit more uniform attention distributions. This observation suggests that attention inadequacy of the instruction arises in the middle layers. Accordingly, in our experimental settings, we opt to incorporate \model into the middle three layers.\par
\begin{figure}[h]
    \centering
    \includegraphics[width=0.9\linewidth]{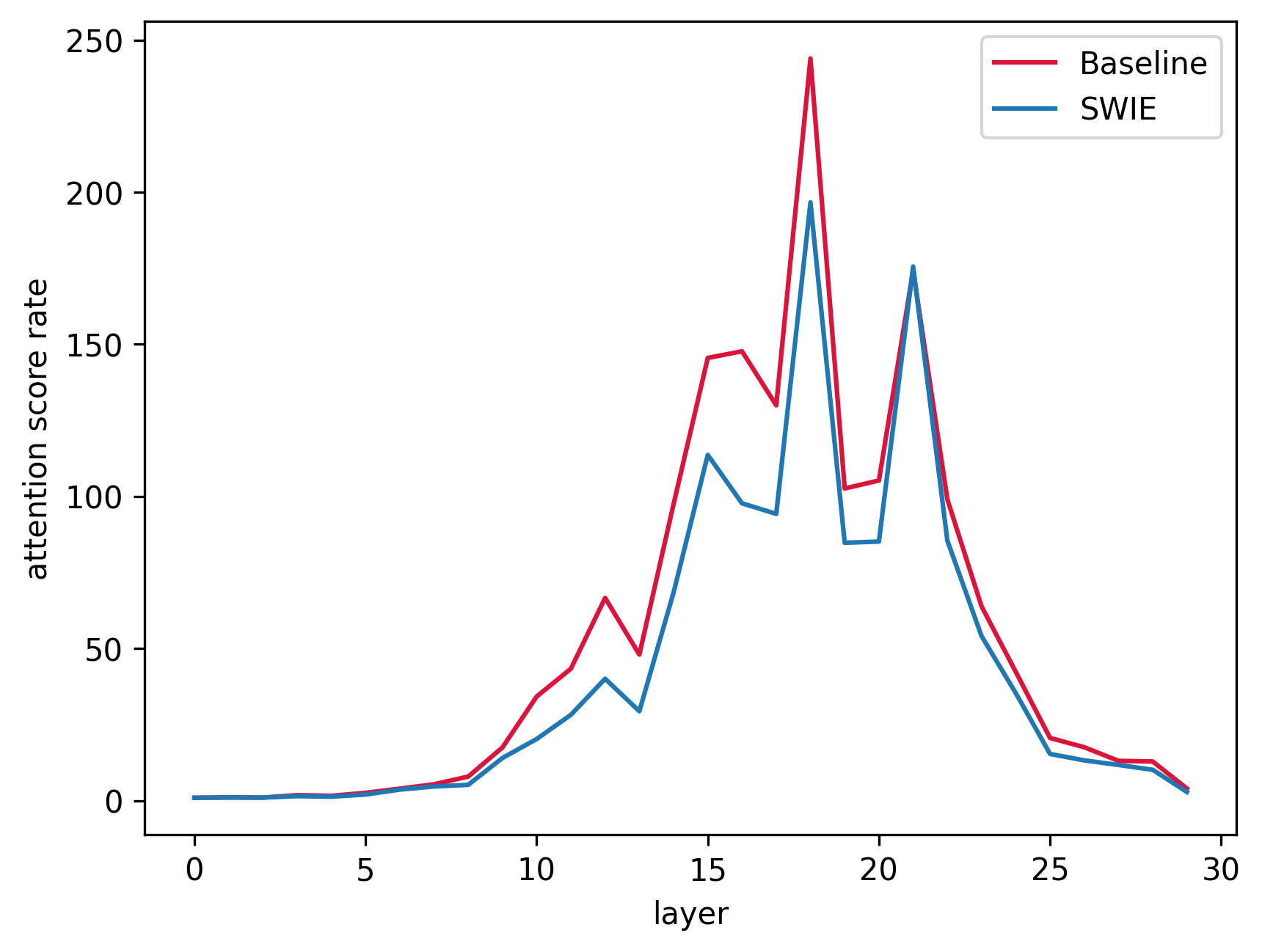}
    \caption{The comparison between models with and without \model on attention ratio between post-instruction and post-input position. This experiment is based on BLOOMZ-3b.}
    \label{fig:attn_rate}
\end{figure}

We compute the ratio of the attention score at the ending position of the instruction and the attention score at the ending position of the input.
As illustrated in Figure.\ref{fig:attn_rate}, our method leads to a lower attention rate, especially for the middle layers, which implies that the attention on the instruction is relatively higher than that of the baseline model.
\begin{equation}
    S_{span} = \sum^{T}_{i=sid+1}a[i][sid]
    \label{eq:accu}
\end{equation}

\section{Conclusion}
We proposed \model and \dataset, a novel additional model structure for strengthening the attention of the model to instruction, and an effective data construction method for machine translation faithfulness. The experiment results show that our methods outperform the strong baselines on widely used machine translation metrics like BLEU and COMET, and \model improves the translation performance more significantly in long text and zero-shot scenarios. To evaluate the translation faithfulness, we employ a cross-lingual word alignment metric, and the result further illustrates the effectiveness of our method on faithful translation. Through the internal attention scores of the models, we visualize the attention distribution on the original model and the attention shift induced by \model, thereby corroborating our assumption regarding the necessity for increased attention on instruction.\par
In the future, the following aspects can be explored based on our work: (1) investigating explainable and trainable methodologies for constructing segment weights; (2) extending the data construction method to other tasks; (3) exploring methods to reduce inference latency.

\bibliography{anthology,custom}

\begin{thebibliography}{25}
\expandafter\ifx\csname natexlab\endcsname\relax\def\natexlab#1{#1}\fi

\bibitem[{Bahdanau et~al.(2015)Bahdanau, Cho, and Bengio}]{bahadanau2015neural}
Dzmitry Bahdanau, Kyunghyun Cho, and Yoshua Bengio. 2015.
\newblock \href {http://arxiv.org/abs/1409.0473} {Neural machine translation by
  jointly learning to align and translate}.

\bibitem[{Choi et~al.(2022)Choi, Jo, Jang, and Seo}]{choi2022prompt}
Eunbi Choi, Yongrae Jo, Joel Jang, and Minjoon Seo. 2022.
\newblock Prompt injection: Parameterization of fixed inputs.
\newblock \emph{arXiv preprint arXiv:2206.11349}.

\bibitem[{Dale et~al.(2023)Dale, Voita, Lam, Hansanti, Ropers, Kalbassi, Gao,
  Barrault, and Costa-juss{\`a}}]{dale2023halomi}
David Dale, Elena Voita, Janice Lam, Prangthip Hansanti, Christophe Ropers,
  Elahe Kalbassi, Cynthia Gao, Lo{\"\i}c Barrault, and Marta~R Costa-juss{\`a}.
  2023.
\newblock Halomi: A manually annotated benchmark for multilingual hallucination
  and omission detection in machine translation.
\newblock \emph{arXiv preprint arXiv:2305.11746}.

\bibitem[{Ferrando et~al.(2022)Ferrando, G{\'a}llego, Alastruey, Escolano, and
  Costa-juss{\`a}}]{ferrando2022towards}
Javier Ferrando, Gerard~I G{\'a}llego, Belen Alastruey, Carlos Escolano, and
  Marta~R Costa-juss{\`a}. 2022.
\newblock Towards opening the black box of neural machine translation: Source
  and target interpretations of the transformer.
\newblock In \emph{Proceedings of the 2022 Conference on Empirical Methods in
  Natural Language Processing}, pages 8756--8769.

\bibitem[{Gao et~al.(2022)Gao, Herold, Yang, and Ney}]{gao2022encoder}
Yingbo Gao, Christian Herold, Zijian Yang, and Hermann Ney. 2022.
\newblock Is encoder-decoder redundant for neural machine translation?
\newblock In \emph{Proceedings of the 2nd Conference of the Asia-Pacific
  Chapter of the Association for Computational Linguistics and the 12th
  International Joint Conference on Natural Language Processing}, pages
  562--574.

\bibitem[{He et~al.(2023)He, Liang, Jiao, Zhang, Yang, Wang, Tu, Shi, and
  Wang}]{he2023exploring}
Zhiwei He, Tian Liang, Wenxiang Jiao, Zhuosheng Zhang, Yujiu Yang, Rui Wang,
  Zhaopeng Tu, Shuming Shi, and Xing Wang. 2023.
\newblock Exploring human-like translation strategy with large language models.
\newblock \emph{arXiv preprint arXiv:2305.04118}.

\bibitem[{Hendy et~al.(2023)Hendy, Abdelrehim, Sharaf, Raunak, Gabr,
  Matsushita, Kim, Afify, and Awadalla}]{hendy2023good}
Amr Hendy, Mohamed Abdelrehim, Amr Sharaf, Vikas Raunak, Mohamed Gabr, Hitokazu
  Matsushita, Young~Jin Kim, Mohamed Afify, and Hany~Hassan Awadalla. 2023.
\newblock How good are gpt models at machine translation? a comprehensive
  evaluation.
\newblock \emph{arXiv preprint arXiv:2302.09210}.

\bibitem[{Honovich et~al.(2022)Honovich, Scialom, Levy, and
  Schick}]{honovich2022unnatural}
Or~Honovich, Thomas Scialom, Omer Levy, and Timo Schick. 2022.
\newblock Unnatural instructions: Tuning language models with (almost) no human
  labor.
\newblock \emph{arXiv preprint arXiv:2212.09689}.

\bibitem[{Jiao et~al.(2023{\natexlab{a}})Jiao, Huang, Wang, Wang, Shi, and
  Tu}]{jiao2023parrot}
Wenxiang Jiao, Jen-tse Huang, Wenxuan Wang, Xing Wang, Shuming Shi, and
  Zhaopeng Tu. 2023{\natexlab{a}}.
\newblock Parrot: Translating during chat using large language models.
\newblock \emph{arXiv preprint arXiv:2304.02426}.

\bibitem[{Jiao et~al.(2023{\natexlab{b}})Jiao, Wang, Huang, Wang, and
  Tu}]{jiao2023chatgpt}
Wenxiang Jiao, Wenxuan Wang, JT~Huang, Xing Wang, and ZP~Tu.
  2023{\natexlab{b}}.
\newblock Is chatgpt a good translator? yes with gpt-4 as the engine.
\newblock \emph{arXiv preprint arXiv:2301.08745}.

\bibitem[{Lee et~al.(2018)Lee, Firat, Agarwal, Fannjiang, and
  Sussillo}]{leehallucinations}
Katherine Lee, Orhan Firat, Ashish Agarwal, Clara Fannjiang, and David
  Sussillo. 2018.
\newblock Hallucinations in neural machine translation.

\bibitem[{Liu et~al.(2023)Liu, Lin, Hewitt, Paranjape, Bevilacqua, Petroni, and
  Liang}]{liu2023lost}
Nelson~F Liu, Kevin Lin, John Hewitt, Ashwin Paranjape, Michele Bevilacqua,
  Fabio Petroni, and Percy Liang. 2023.
\newblock Lost in the middle: How language models use long contexts.
\newblock \emph{arXiv preprint arXiv:2307.03172}.

\bibitem[{Longpre et~al.(2023)Longpre, Hou, Vu, Webson, Chung, Tay, Zhou, Le,
  Zoph, Wei et~al.}]{longpre2023flan}
Shayne Longpre, Le~Hou, Tu~Vu, Albert Webson, Hyung~Won Chung, Yi~Tay, Denny
  Zhou, Quoc~V Le, Barret Zoph, Jason Wei, et~al. 2023.
\newblock The flan collection: Designing data and methods for effective
  instruction tuning.
\newblock \emph{arXiv preprint arXiv:2301.13688}.

\bibitem[{M{\"u}ller et~al.(2020)M{\"u}ller, Rios, and
  Sennrich}]{muller-2020-domain}
Mathias M{\"u}ller, Annette Rios, and Rico Sennrich. 2020.
\newblock Domain robustness in neural machine translation.
\newblock pages 151--164.

\bibitem[{Papineni et~al.(2002)Papineni, Roukos, Ward, and
  Zhu}]{papineni2002bleu}
Kishore Papineni, Salim Roukos, Todd Ward, and Wei-Jing Zhu. 2002.
\newblock Bleu: a method for automatic evaluation of machine translation.
\newblock In \emph{Proceedings of the 40th annual meeting of the Association
  for Computational Linguistics}, pages 311--318.

\bibitem[{Raffel et~al.(2020)Raffel, Shazeer, Roberts, Lee, Narang, Matena,
  Zhou, Li, and Liu}]{raffel2020exploring}
Colin Raffel, Noam Shazeer, Adam Roberts, Katherine Lee, Sharan Narang, Michael
  Matena, Yanqi Zhou, Wei Li, and Peter~J Liu. 2020.
\newblock Exploring the limits of transfer learning with a unified text-to-text
  transformer.
\newblock \emph{The Journal of Machine Learning Research}, 21(1):5485--5551.

\bibitem[{Raunak et~al.(2021)Raunak, Menezes, and
  Junczys-Dowmunt}]{raunak2021curious}
Vikas Raunak, Arul Menezes, and Marcin Junczys-Dowmunt. 2021.
\newblock The curious case of hallucinations in neural machine translation.
\newblock In \emph{Proceedings of the 2021 Conference of the North American
  Chapter of the Association for Computational Linguistics: Human Language
  Technologies}, pages 1172--1183.

\bibitem[{Raunak et~al.(2023)Raunak, Sharaf, Awadallah, and
  Menezes}]{raunak2023leveraging}
Vikas Raunak, Amr Sharaf, Hany~Hassan Awadallah, and Arul Menezes. 2023.
\newblock Leveraging gpt-4 for automatic translation post-editing.
\newblock \emph{arXiv preprint arXiv:2305.14878}.

\bibitem[{Wang et~al.(2022)Wang, Mishra, Alipoormolabashi, Kordi, Mirzaei,
  Naik, Ashok, Dhanasekaran, Arunkumar, Stap et~al.}]{wang2022super}
Yizhong Wang, Swaroop Mishra, Pegah Alipoormolabashi, Yeganeh Kordi, Amirreza
  Mirzaei, Atharva Naik, Arjun Ashok, Arut~Selvan Dhanasekaran, Anjana
  Arunkumar, David Stap, et~al. 2022.
\newblock Super-naturalinstructions: Generalization via declarative
  instructions on 1600+ nlp tasks.
\newblock pages 5085--5109.

\bibitem[{Wei et~al.(2021)Wei, Bosma, Zhao, Guu, Yu, Lester, Du, Dai, and
  Le}]{wei2021finetuned}
Jason Wei, Maarten Bosma, Vincent Zhao, Kelvin Guu, Adams~Wei Yu, Brian Lester,
  Nan Du, Andrew~M Dai, and Quoc~V Le. 2021.
\newblock Finetuned language models are zero-shot learners.

\bibitem[{Wei et~al.(2022)Wei, Tay, Bommasani, Raffel, Zoph, Borgeaud,
  Yogatama, Bosma, Zhou, Metzler et~al.}]{wei2022emergent}
Jason Wei, Yi~Tay, Rishi Bommasani, Colin Raffel, Barret Zoph, Sebastian
  Borgeaud, Dani Yogatama, Maarten Bosma, Denny Zhou, Donald Metzler, et~al.
  2022.
\newblock Emergent abilities of large language models.
\newblock \emph{arXiv preprint arXiv:2206.07682}.

\bibitem[{Ye et~al.(2022)Ye, Kim, Jang, Shin, and Seo}]{ye2022guess}
Seonghyeon Ye, Doyoung Kim, Joel Jang, Joongbo Shin, and Minjoon Seo. 2022.
\newblock Guess the instruction! flipped learning makes language models
  stronger zero-shot learners.
\newblock In \emph{The Eleventh International Conference on Learning
  Representations}.

\bibitem[{Zeng et~al.(2023)Zeng, Meng, Yin, and Zhou}]{zeng2023tim}
Jiali Zeng, Fandong Meng, Yongjing Yin, and Jie Zhou. 2023.
\newblock Tim: Teaching large language models to translate with comparison.
\newblock \emph{arXiv preprint arXiv:2307.04408}.

\bibitem[{Zhang et~al.(2023)Zhang, Fang, Zhang, Ma, Zhou, Huang, Bu, Gui, Chen,
  Chen et~al.}]{zhang2023bayling}
Shaolei Zhang, Qingkai Fang, Zhuocheng Zhang, Zhengrui Ma, Yan Zhou, Langlin
  Huang, Mengyu Bu, Shangtong Gui, Yunji Chen, Xilin Chen, et~al. 2023.
\newblock Bayling: Bridging cross-lingual alignment and instruction following
  through interactive translation for large language models.
\newblock \emph{arXiv preprint arXiv:2306.10968}.

\bibitem[{Zhao et~al.(2023)Zhao, Zhou, Li, Tang, Wang, Hou, Min, Zhang, Zhang,
  Dong et~al.}]{zhao2023survey}
Wayne~Xin Zhao, Kun Zhou, Junyi Li, Tianyi Tang, Xiaolei Wang, Yupeng Hou,
  Yingqian Min, Beichen Zhang, Junjie Zhang, Zican Dong, et~al. 2023.
\newblock A survey of large language models.
\newblock \emph{arXiv preprint arXiv:2303.18223}.

\end{thebibliography}
\bibliographystyle{acl_natbib}




\end{document}